\documentclass[times, review, 10pt]{elsarticle}

\usepackage{amssymb}
\usepackage{amsmath}
\usepackage{adjustbox}
\usepackage{booktabs}
\usepackage{chngcntr}
\usepackage{etoolbox} %

\usepackage{cleveref}
\Crefname{figure}{Fig.}{Figures}
\usepackage{chngcntr}
\usepackage{cleveref}
\counterwithout{table}{section}            %
\counterwithout{table}{section}

\crefname{table}{Table}{Tables}  %
\Crefname{table}{Table}{Tables}

\journal{Pattern Recognition}
\usepackage{xcolor} 
\usepackage{colortbl}
\usepackage{multirow}
\usepackage{graphicx}
\usepackage{tabularx}
\usepackage{bm}
\usepackage{chngcntr}

\def\eg{\emph{e.g.}}

\newcommand{\tr}[1]{{\color{red} #1}}
\def\tsc#1{\csdef{#1}{\textsc{\lowercase{#1}}\xspace}}
\tsc{WGM}
\tsc{QE}
\begin{document}

\begin{frontmatter}
\title{LDFE: Laplacian Decoupled Feature Enhancement Block for Dual-Stream CNN-based RGB-IR Object Detection}

\author[1]{\textcolor{black}{Wenhao} Dong}

\ead{ZB2315207@buaa.edu.cn}

\affiliation[1]{
            addressline={Beihang University}, 
            city={Beijing},
            postcode={102206}, 
            country={China}}

\author[2,3]{\textcolor{black}{Xiaoyan} Luo}%
\affiliation[2]{organization={State Key Laboratory of High-Efficiency Reusable Aerospace Transportation Technology, The School of Astronautics},
            addressline={Beihang University}, 
            city={Beijing},
            postcode={102206}, 
            country={China}}
\affiliation[3]{organization={The National Key Laboratory of Integrated Air-Ground Navigation Technologies, The School of Astronautics},
            addressline={Beihang University}, 
            city={Beijing},
            postcode={102206}, 
            country={China}}
\author[4]{\textcolor{black}{Linlin} Yang\corref{cor1}}%
\ead{lyang@cuc.edu.cn}

\affiliation[4]{organization={The State Key Laboratory of Media Convergence and Communication},
            addressline={Communication University of China}, 
            city={Beijing},
            postcode={100024}, 
            country={China}}
\author[5,6]{\textcolor{black}{Haodong} Zhu}

\affiliation[5]{organization={The School of Artificial Intelligence},
            addressline={Beihang University}, 
            city={Beijing},
            postcode={100191}, 
            country={China}}
\affiliation[6]{organization={Zhongguancun Academy},
            city={Beijing},
            postcode={100094}, 
            country={China}}
\author[7,8]{\textcolor{black}{Xiaorong} Shi}
\affiliation[7]{organization={The School of Automation Science and Electrical Engineering},
            addressline={Beihang University}, 
            city={Beijing},
            postcode={100091}, 
            country={China}}
\affiliation[8]{organization={Beijing Institute of Control and Electronics Technology},
            city={Beijing},
            postcode={100038}, 
            country={China}}

\author[9]{\textcolor{black}{Guodong} Guo}

\affiliation[9]{organization={Ningbo Institute of Digital Twin},
            addressline={Eastern Institute of Technology}, 
            city={Ningbo},
            postcode={315200}, 
            state={Zhejiang},
            country={China}}
\author[5,10]{\textcolor{black}{Baochang} Zhang}

\affiliation[10]{organization={Hangzhou Research Institute, School of Artificial Intelligence},
            addressline={Beihang University}, 
            city={Hangzhou},
            postcode={310020}, 
            state={Zhejiang},
            country={China}}

\cortext[cor1]{Corresponding author} 
\begin{abstract}
The complementary information between RGB and IR images can significantly enhance object detection performance under extreme conditions. 
Existing methods prefer dual-stream CNN backbones built upon YOLO for feature extraction and focus on the design of feature fusion.
{In this paper, we introduce the {Laplacian Decoupled Feature Enhancement block} (LDFE) to fuse features from different stages of the dual-stream CNN {backbone}. 
By design, LDFE simultaneously considers the characteristics of modalities and structures for feature fusion by employing global-local decomposition, denoising, fusion, and reconstruction, sequentially.
The LDFE first separates features into global and local components based on Laplacian Pyramid, and then performs denoising and fusion based on Global State Space Enhancement module (GS$^2$E) and Local Convolutional Correlation Enhancement module (LC$^2$E) separately.
Specifically, the GS$^2$E conducts a two-branch architecture for the main and auxiliary modalities.
It dynamically suppresses noise in the main modality through cross-modal attention derived from the auxiliary modality, while employing a State Space Model to capture long-range dependencies within the global feature representations of the main modality. To obtain bidirectional interaction, the two modalities systematically alternate their main/auxiliary roles. 
Moreover, the LC$^2$E suppresses noise in local features and leverages spatial and channel dimension along with triple convolution to extract fine-grained details for fusion.}
These innovative designs achieve a significant performance improvement, with \bm{$m$}AP surpassing the SOTA methods \(\text{6.2\%}\), \(\text{3.7\%}\), \(\text{4.7\%}\), \(\text{2.3\%}\), \(\text{4.1\%}\) and \(\text{2.0\%}\) on M$^3$FD, DroneVehicle, LLVIP, FLIR-Aligned, KAIST and VEDAI datasets, respectively.
\end{abstract}

\end{frontmatter}
\section{Introduction}
\label{sec:intro}

Deep-learning based object detection holds vast application potential in areas such as road surveillance and autonomous driving~\cite{sun2024mrd}. However, under extreme conditions such as low illumination and adverse weather, traditional detection methods that rely on a single modality often exhibit limited effectiveness~\cite{chen2023igt}. %
RGB images can provide clear object details and texture information, but are highly susceptible to imaging conditions. In contrast, IR images are less affected by imaging conditions and can offer clear object contour information but they lack fine object details~\cite{shen2024icafusion,zhang2024e2e,li2025reference}.
With the advancement of multimodal sensing techniques, utilizing paired RGB-IR images for detection has emerged as a new development direction in the field of object detection as they can provide complementary information to each other. 
{Due to the strong performance and lightweight structure, the dual-stream CNN architecture built upon YOLO~\cite{fei2024acdf,tang2024yolo,chen2024rgb,tian2024iv} has become one of the mainstream backbones for RGB-IR object detection, and existing works focus on the fusion of RGB and IR features of different streams.}

{Architecture-wise, existing fusion designs for dual-stream networks
explore CNN~\cite{zhao2024removal,wang2025haarfuse,zhao2025differential}, Transformer~\cite{rs16203904,10877920,10570450}, or Mamba~\cite{dong2025fusion,zhou2024dmm} to directly process features from the two modalities, achieving significant improvements in detection performance. 
{However, different structures emphasize different aspects and also possess distinct limitations when fusing features.} For example, CNN excels at capturing local spatial features, but is limited by receptive fields, struggling to capture global information~\cite{cordonnier2020relationship}.
In contrast, {Transformers} and Mamba are proficient in modeling global information and long-range dependencies but often exhibit a deficiency in their sensitivity to fine-grained details~\cite{ma2024tinyvim}. 
{Given these structural limitations—where no single structure perfectly balances global and local modeling—it is critical to revisit the inherent characteristics of the input modalities themselves. Specifically, RGB and IR inherently emphasize distinct aspects of {global and local details} with their own modality noise. This modal dichotomy, local vs. global, naturally aligns with the complementary strengths of CNN (local) and Transformer/Mamba (global).
{This prompts us to apply a global-local decomposition and corresponding noise suppression within the fusion process, while aligning the design with the intrinsic attributes of network structures—ultimately enabling more comprehensive and efficient feature fusion.}
For decomposition, the Laplacian Pyramid is a simple yet effective method for multi-scale feature decomposition. At a given scale, it can separate features into global and local components, effectively preserving fine-grained information while also enabling the original features to be reconstructed through straightforward operations. Therefore, integrating the dual-stream CNN-based YOLO framework with the Laplacian Pyramid-based decomposition, and designing appropriate fusion structures for decomposed global-local features, offers a compelling and innovative strategy to further improve detection performance, as shown in Fig.~\ref{fig:intro}.}

{In this paper, we propose a novel fusion block, Laplacian Decoupled Feature Enhancement (LDFE) as shown in Fig.~\ref{fig:actitecture}. LDFE is a four-step paradigm that includes Laplacian decomposition, denoising, fusion, and the final Laplacian reconstruction, which is applied for different {stages} of the dual-stream CNN {backbone}.
For decomposition, features are used as the base of the multi-scale pyramid in LDFE. The Gaussian-smoothed feature and the first layer of the Laplacian Pyramid are utilized as global and detail features, respectively, at the current scale.}
{For denoising and fusion, we design the Global State Space Enhancement module (GS$^2$E) and the Local Convolutional Correlation Enhancement module (LC$^2$E) to implement information extraction and fusion on global and detail features, respectively. As inherent noise in RGB and IR features can also affect the performance of the detector~\cite{xiong2025efficient}, we also integrate a denoising process into our modules.}
The final output is derived through the integration of global and local features using Laplacian reconstruction.
Unlike Fusion-Mamba~\cite{dong2025fusion} utilizing the dual-stream Mamba to suppress noise holistically, we adopt tailored denoising strategies for global and local features based on the distinct characteristics of noise in each. 
In GS$^2$E, after exchanging half of the channels, the features processed by successive convolutions of one modality are used as attention to suppress local noise in the other modality and features of the two modalities take turns serving as attention. Subsequently, the residual Mamba structure is employed to extract global information. In LC$^2$E, we employ the $L_{1}$ Normalization to suppress noisy pixels in the local features of both modalities. Subsequently, Softmax Fusion is applied to integrate the features, followed by spatial and channel attention to fully extract detailed information. 
\begin{figure}[t]
  \centering
   \includegraphics[width=0.9\linewidth]{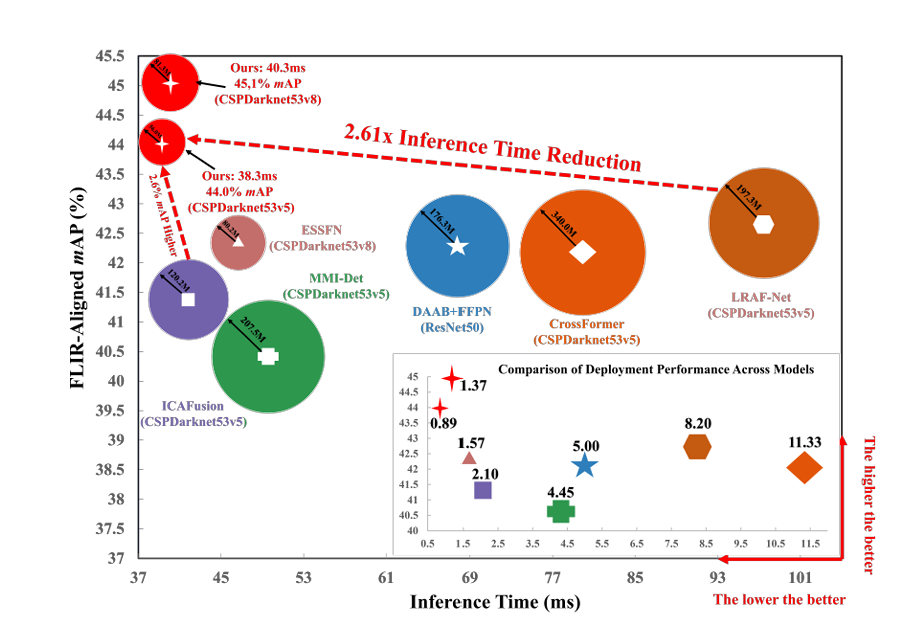}
   \caption{{Comparison of our model with several multimodal detection methods on FLIR-Aligned dataset, evaluating $m$AP, Inference Time, and Parameters. The size of the circles represents the parameters. {Here, the deployment performance is evaluated by dividing the product of parameters and inference time by the typical deployment requirement for mobile devices ($60$M × $40$ms). A lower ratio indicates better compliance with deployment constraints.} Our model achieves the best performance.}}
   \label{fig:intro}
\end{figure}

{The integration of the dual-stream CNN backbones with LDFE significantly improves the global and local information representation capability of multi-scale features, leading to a substantial enhancement in detection performance. Thanks to the lightweight CNN backbone and LDFE, our proposed framework demonstrates both high efficiency and effectiveness.} As shown in Fig.~\ref{fig:intro}, on the challenging dataset FLIR-Aligned, our method with two kinds of CNN-based backbones
significantly outperforms the current SOTA approaches in terms of $m$AP, F1 Score, and other metrics. Benefiting from Mamba's linear complexity, our method also achieves optimal performance in both parameter efficiency and runtime. In summary, the contributions of this paper are as follows:
\begin{itemize}
    \item{We propose Laplacian Decoupled Feature Enhancement (LDFE). By design, it simultaneously considers the characteristics of modalities and structures for feature fusion by employing global-local decomposition, denoising, fusion, and reconstruction, sequentially.
    }
    {\item The LDFE comprises two sub-modules, each thoughtfully designed to separately denoise and fuse global and local features of RGB and IR: the GS$^2$E, which realizes effective global feature fusion through dual-modal alternating attention and a state-space model, and the LC$^2$E, which enhances the integration of local detail information through an innovative attention mechanism.}
    \item Extensive experiments on six RGB-IR object detection datasets and three typical CNN backbones demonstrate the effectiveness of our method. Compared to the second-best method, our approach achieves an average $\textcolor{black}{m}$AP improvement of $3.83$\%.
\end{itemize}
\begin{figure*}[!t]
  \centering
   \includegraphics[width=0.9\linewidth]{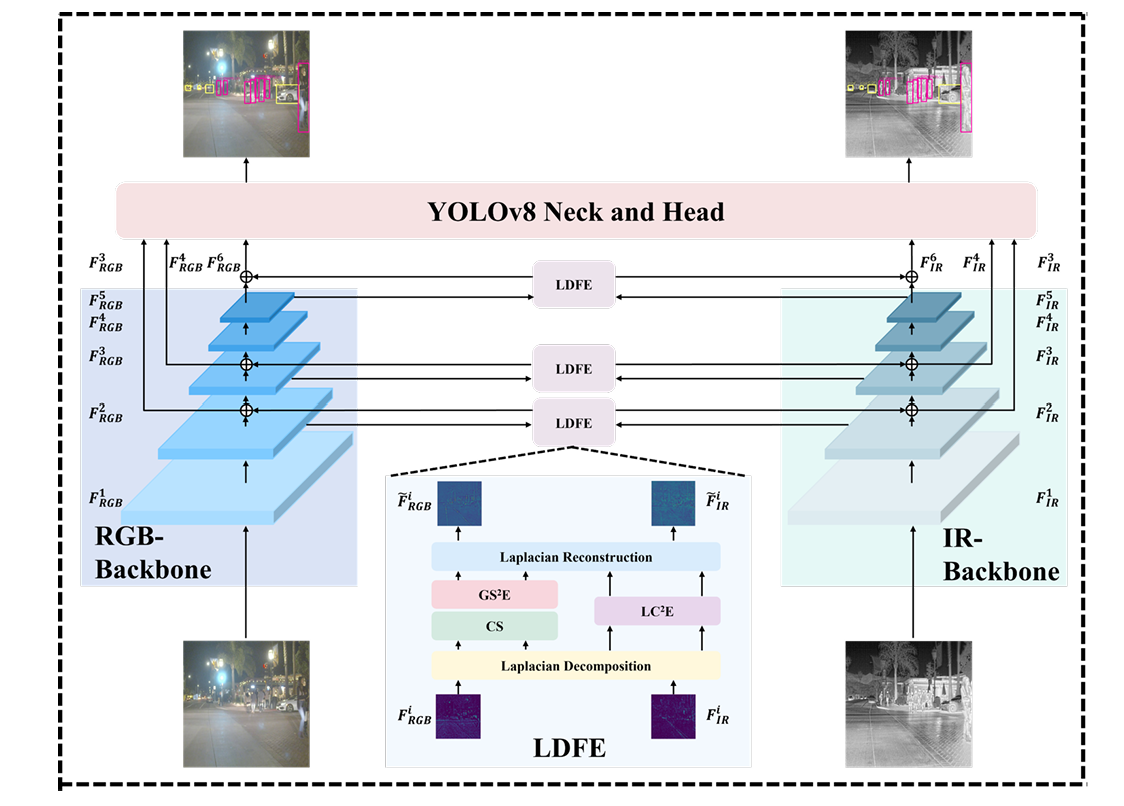}
   \caption{{An illustrative diagram of our proposed method, which consists of a dual-stream CNN architecture built upon YOLO with LDFEs at layers $2$, $3$, and $5$. Specifically, LDFE comprises Laplacian Decomposition for global and local feature decomposition, Channel Swapping (CS) for sufficient global feature interaction, Global State Space Enhancement module (GS$^{2}$E) for global feature denoising and fusion, Local Convolutional Correlation Enhancement module (LC$^{2}$E) for local feature denoising and fusion, and the final Laplacian Reconstruction.}}
   \label{fig:actitecture}
\end{figure*}

\section{Related Works}
\label{sec:related}
\subsection{Multimodal Object Detection}
With the rapid development of object detectors~\cite{yolov8_ultralytics}, many multimodal detection methods that leverage complementary information from multiple spectra have been proposed. {The key to these methods lies in the design of the fusion strategy, which can be categorized into pixel-level fusion methods~\cite{tang2023divfusion,ding2025fiafusion} and feature-level fusion methods~\cite{yang2024multidimensional,sun2024adaptive,xu2024enhanced}.}  Pixel-level {fusion 
} methods fuse the input RGB and IR image pairs at the pixel level to create a single fused image, which is then used for object detection, such as EMMA~\cite{zhao2024equivariant} and CoCoNet~\cite{liu2024coconet}. Feature-level {fusion} methods fuse the RGB and IR information at the feature level. Based on the fusion stage of the features, they can be classified into early and middle~\cite{lee2024crossformer,yuan2024c,zhou2024vehicle,10144688} (fusing early or later features from the dual-stream backbones) and late fusion~\cite{guo2024damsdet,wang2024kcdnet} (fusing the outputs from independent detectors). Most of these fusion methods are based on CNN or Transformer. {Recently, some Mamba-based methods such as Fusion-Mamba~\cite{dong2025fusion} and DMM~\cite{zhou2024dmm}, have leveraged Mamba's capability of linearly modeling global features to achieve further performance improvement in multimodal object detection tasks.} {In this paper, we highlight the importance of separately denoising and fusing multimodal global structural information and local detail information within multi-scale features, and design fusion strategies GS$^2$E and LC$^2$E, which can further enhance detection performance.} %
\subsection{Laplacian Pyramid Methods in Computer Vision}
The Laplacian Pyramid can decompose multi-scale {global features} and capture the overall structure while preserving local features that emphasize edges and textures. Methods based on the Laplacian Pyramid are widely used in various computer vision tasks with great performance. {LapH~\cite{luo2023multi} decomposes multi-scale high-frequency and low-frequency components using the Laplacian Pyramid, and guides the feature extraction of low-frequency components with high-frequency component information, thereby improving the quality of infrared and visible image fusion.} LPSR~\cite{zhang2022multispectral} also uses the Laplacian Pyramid to extract high-frequency and low-frequency features, which are then sparsely represented. Through the selective fusion of these features, it establishes a new SOTA benchmark in multi-spectral and SAR image fusion. %
{These methods have been successful in extracting both global and local features across multiple scales. Given its importance in multimodal feature fusion, we further develop this idea by introducing LDFE, which emphasizes utilizing the Laplacian Pyramid for feature fusion of multimodal object detection tasks.}%

{\section{Preliminaries}}
The Laplacian Pyramid is a multi-scale image representation method constructed based on the Gaussian Pyramid~\cite{burt1987laplacian}. By decomposing the image into multiple scales using the Gaussian Pyramid, it separates the global feature while preserving fine local details of the image. 
{Specifically, the Laplacian Pyramid can be treated as a combination of a series of decomposition and reconstruction.
For each scale, the decomposition operation aims to decouple global and local features at the current scale, while the reconstruction operation generates features transmitted to the next layer.}

For \textbf{decomposition}, given initial features ${F}_{I}$, its global feature $F_G$ is obtained via Gaussian Smoothing $\text{GS}(\cdot)$, which effectively suppress edge and detail information within the features. 
Consequently, its local features $F_L$ at the current scale are computed by subtracting the downsampled and then upsampled result of $F_G$ from the initial features ${F}_{I}$:
\begin{equation}
\begin{aligned}
&{F}_{G}=\text{GS}({F}_{I}),\\
&{F}_{L}={F}_{I}-\text{UpSample}(\text{DownSample(${F}_{G}$)}),
\end{aligned}
\label{eq:2}
\tag{1}
\end{equation}
where $\text{UpSample}(\cdot)$ and $\text{DownSample}(\cdot)$ denote the upsampling and downsampling operators.

For \textbf{reconstruction}, the processed global and local features $F_G$ and $F_L$ are summed to obtain the reconstructed features ${F}_{R}$ to be propagated to the subsequent layers as:
\begin{equation}
\begin{aligned}
&{F}_{R}={F}_{G}+{F}_{L}.\\
\end{aligned}
\label{eq:3}
\tag{2}
\end{equation}\par

\section{Architecture}
The overall architecture of our model is illustrated in Fig.~\ref{fig:actitecture}, which consists of a dual-stream CNN feature extraction backbone, three LDFE modules, and the YOLOv8 detection neck and head. Given paired visible $I_{RGB}$ and infrared images $I_{IR}$, our target is to leverage the complementary information of paired multimodal images to detect target objects in the images, providing both the categories and the bounding boxes.
{We employ the commonly used five-stage CNN backbones (\eg, ResNet50), and extend them into dual-stream architecture to process paired image inputs. {For each stage in the backbone, we use its output as the base level of Gaussian Pyramid.} Especially, we place LDFEs at the $2$nd, $3$rd, and $5$th stages to fuse IR and RGB features.}{ The processed features from these three stages are summed and subsequently fed into the original YOLOv8 neck and head to get detection results.}
We adopt the standard YOLOv8 loss function~\cite{yolov8_ultralytics} formally expressed as:
\begin{equation}
L_{detection}=\lambda_{coord}L_{coord}+L_{conf}+L_{class},
\tag{4}
\end{equation}
where $L_{coord}$ represents the localization loss, $\lambda_{coord}$ is the corresponding hyperparameter, $L_{conf}$ denotes the confidence loss, and $L_{class}$ represents the classification loss. More details of the loss function can be found in~\cite{yolov8_ultralytics}.

\subsection{Dual-Stream Backbone}
Our model first extracts features from $I_{RGB}$ and $I_{IR}$ through the first two {stages} of the dual-stream backbone, obtaining $F^{i}_{RGB}$ and $F^{i}_{IR}$, where $i\in\{1,2\}$. 
Subsequently, $F^{2}_{RGB}$ and $F^{2}_{IR}$ are fed into the LDFE and result in $\tilde{F}^{2}_{RGB}$ and $\tilde{F}^{2}_{IR}$. Specifically, during LDFE, $F^{2}_{RGB}$ and $F^{2}_{IR}$ are used as initial features for decomposition (See Eq.~\ref{eq:2}). After denoising, 
fusion, and feature extraction in LDFE,  the reconstructed features $\tilde{F}^{2}_{RGB}$ and $\tilde{F}^{2}_{IR}$ are obtained (See Eq.~\ref{eq:3}).
The sum of $F^{2}_{RGB}$ and $\tilde{F}^{2}_{RGB}$ and the sum of $F^{2}_{IR}$ and $\tilde{F}^{2}_{IR}$ are sent to the third {stage} of the backbone, yielding $F^{3}_{RGB}$ and $F^{3}_{IR}$, respectively. 
To obtain more multi-scale feature information combining the Laplacian Pyramid and the dual-stream backbone, we also apply {LDFE} in the third and fifth {stages} of the backbone as : 
\begin{equation}
\begin{aligned}
&{F}^{i+1}_{RGB}=\tilde{F}^{i}_{RGB}+{F}^{i}_{RGB},\\
&{F}^{i+1}_{IR}=\tilde{F}^{i}_{IR}+{F}^{i}_{RGB},
\end{aligned}
\tag{5}
\end{equation}
where $i\in\{2,3,5\}$.
Finally, {the values ${F}^{i+1}_{RGB}$ and ${F}^{i+1}_{IR}$ are summed for $i \in \{2,3,5\}$,} and then fed into the YOLOv8 neck and head to obtain the final detection results. 
For the fourth layer, we employ the original backbone to process ${F}^{4}_{RGB}$ and ${F}^{4}_{IR}$, obtaining ${F}^{5}_{RGB}$ and ${F}^{5}_{IR}$. 
{For the simplicity, we denote the sum of $\tilde{F}^{5}_{RGB}$ and ${F}^{5}_{RGB}$ as ${F}^{6}_{RGB}$ and the sum of $\tilde{F}^{5}_{IR}$ and ${F}^{5}_{IR}$ as ${F}^{6}_{IR}$. Along with ${F}^{3}_{RGB}$, ${F}^{3}_{IR}$, ${F}^{4}_{RGB}$, and ${F}^{4}_{IR}$, they are subsequently delivered to the YOLO Neck and Head to get the detection results.}
We will provide a detailed introduction of LDFE in Section~\ref{ELMCB}.
\subsection{Laplacian Decoupled Feature Enhancement block}
\label{ELMCB}
{To achieve more effective denoising and process both low-frequency global information and high-frequency detail information more effectively at current scale, we design LDFE, as illustrated in Fig.~\ref{fig:actitecture}.} We obtain two sub-features {through Eq.~\ref{eq:2}} from $F^{i}_{RGB}$, where the global sub-feature is retained as ${F}^{i}_{G,RGB}$, and the local sub-feature is ${F}^{i}_{L,RGB}$. The same operation is applied to $F^{i}_{IR}$, obtaining ${F}^{i}_{G,IR}$ and ${F}^{i}_{L,IR}$. Subsequently, ${F}^{i}_{G,RGB}$ and ${F}^{i}_{G,IR}$ are fed into Channel Swapping and Global State Space Enhancement module for denoising, feature fusion, and global information extraction, resulting in fused global features $\tilde{F}^{i}_{G,RGB}$ and $\tilde{F}^{i}_{G,IR}$, which can be formulated as follows:
\begin{equation}
\begin{aligned}
&\bar{F}^{i}_{G,RGB},~\bar{F}^{i}_{G,IR}=\text{CS}({F}^{i}_{G,RGB},{F}^{i}_{G,IR}),\\
&\tilde{F}^{i}_{G,RGB},~\tilde{F}^{i}_{G,IR}=\text{GS$^2$E}(\bar{F}^{i}_{G,RGB},\bar{F}^{i}_{G,IR}).
\end{aligned}
\label{eq:7}
\tag{6}
\end{equation}
Here, $\text{CS}(\cdot)$ represents the Channel Swapping operation, which exchanges half of the channels of the RGB and IR global features to achieve initial information interaction. 
$\bar{F}^{i}_{G,RGB}$ and $\bar{F}^{i}_{G,IR}$ are channel swapped global features.
$\text{GS$^2$E}(\cdot)$ represents the Global State Space Enhancement module. 

The local {sub-feature} ${F}^{i}_{L,RGB}$ and ${F}^{i}_{L,IR}$ are processed by Local Convolutional Correlation Enhancement module for denoising fusion and local feature extraction, formulated as:
\begin{equation}
\tilde{F}^{i}_{L,RGB},\tilde{F}^{i}_{L,IR}=\text{LC$^2$E}({F}^{i}_{L,RGB},{F}^{i}_{L,IR}).
\tag{7}
\end{equation}
Here, $\text{LC$^2$E}(\cdot)$ refers to Local Convolutional Correlation Enhancement module. Finally, $\tilde{F}^{i}_{G,RGB}$ and $\tilde{F}^{i}_{L,RGB}$, $\tilde{F}^{i}_{G,IR}$ and $\tilde{F}^{i}_{L,IR}$ are processed by the reconstruction operation through Eq.~\ref{eq:3}, obtaining the fused features $\tilde{F}^{i}_{RGB}$ and $\tilde{F}^{i}_{IR}$, respectively.  \begin{figure}[!t]
   \centering
    \includegraphics[width=0.8\linewidth]{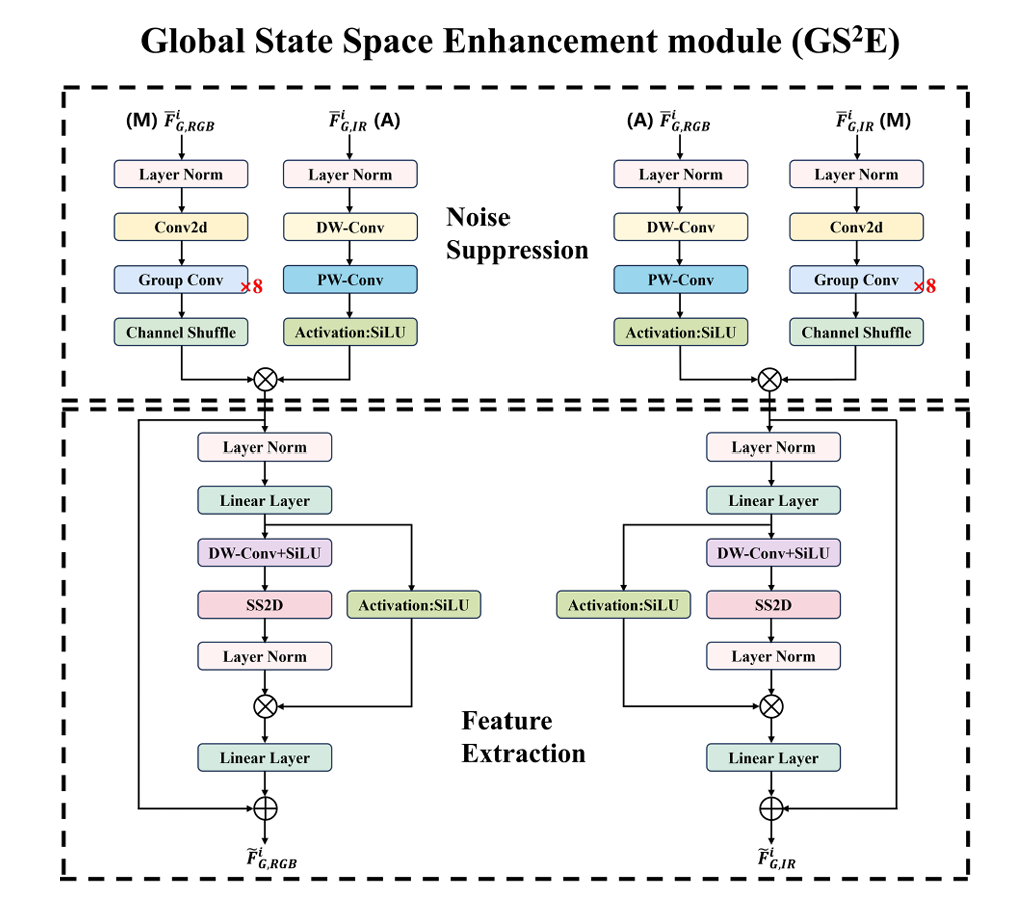}
    \caption{The detailed framework of the Global State Space Enhancement module. {M and A denote Main branch and Auxiliary branch, respectively. }}
    \label{fig:jiegou2}
 \end{figure}

\noindent\textbf{Global State Space Enhancement module (GS$^2$E).}
As shown in Fig.~\ref{fig:jiegou2}, the Global State Space Enhancement module consists of Noise Suppression and Feature Extraction.
{Due to the modality-specific local ``pseudo-label'' noise in infrared and visible features~\cite{dong2025fusion}, we first design the convolution-based mutual suppression in GS$^2$E module.}
{$\bar{F}^{i}_{G,RGB}$ and $\bar{F}^{i}_{G,IR}$ from Eq.~\ref{eq:2} are used as the input of the main {denoised} branch and the auxiliary denoising branch.}
The main branch features undergo normalization, convolution, eight group convolutions, and channel shuffling for feature extraction. Meanwhile, the auxiliary branch undergoes normalization, DepthWise (DW) convolution, PointWise (PW) convolution, and SiLU activation. The resulting features of the auxiliary branch serve as denoising attention, which is multiplied with the main branch features to obtain the denoised output of the main branch. The formula is expressed as follows:
\begin{equation}
\begin{aligned}
&\check{F}^{i}_{G,RGB}=\text{MB}(\bar{F}^{i}_{G,RGB})\odot \text{AB}(\bar{F}^{i}_{G,IR}),\\
&\check{F}^{i}_{G,IR}=\text{MB}(\bar{F}^{i}_{G,IR})\odot \text{AB}(\bar{F}^{i}_{G,RGB}),
\end{aligned}
\tag{8}
\end{equation}
where $\check{F}^{i}_{G,RGB}$ and $\check{F}^{i}_{G,IR}$ represent the denoised global features of RGB and IR, respectively, $\text{MB} (\cdot)$ denotes the main branch operation, $\text{AB} (\cdot)$ represents the auxiliary branch operation, and $\odot$ denotes the Hadamard product. Inspired by~\cite{ma2024tinyvim}, we utilize Mamba to process the denoised fused global feature in feature extraction. {It undergoes normalization, a linear layer, and DW convolution before being fed into the Mamba core module SS$2$D for feature extraction.} Finally, the residual fused features are added to obtain the main branch output. To enable comprehensive bidirectional interaction between features from both modalities, the two modalities alternately serve as the main branch, resulting in the final fused global output $\tilde{F}^{i}_{G,RGB}$ and $\tilde{F}^{i}_{G,IR}$.
 \begin{figure}[!t]
   \centering
    \includegraphics[width=0.95\linewidth]{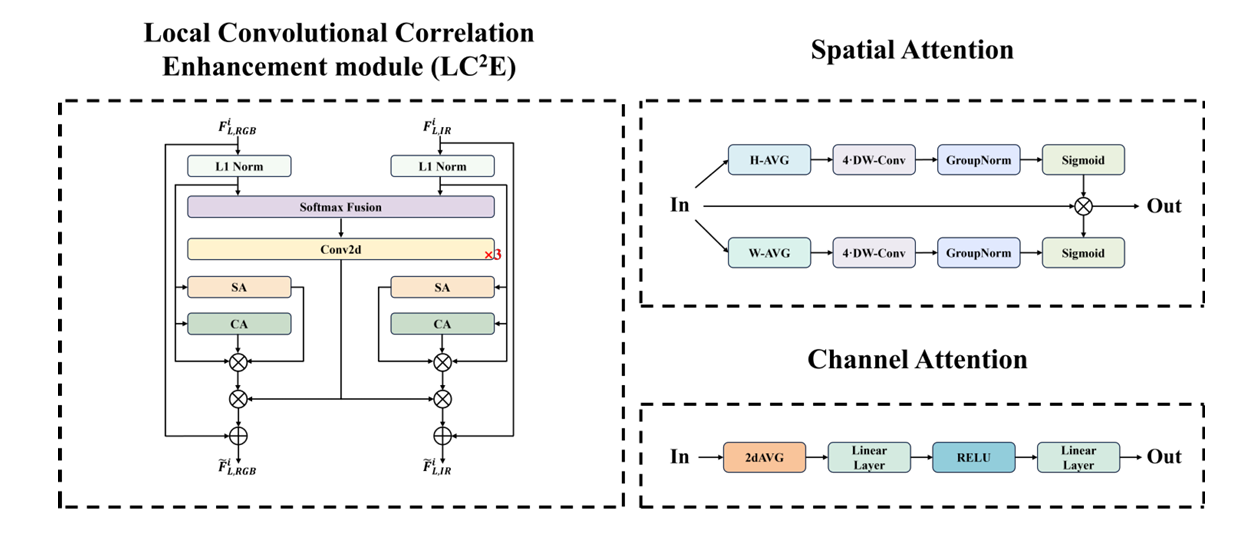}
    \caption{{The detailed structure of the Local Convolutional Correlation Enhancement module with Spatial Attention (SA) and Channel Attention (CA). {L1 Norm means $L_1$ Normalization. }}}
    \label{fig:jiegou3}
 \end{figure}%

\noindent\textbf{{Local Convolutional Correlation Enhancement module} (LC$^2$E).}
The LC$^2$E is illustrated at the {top} of {Fig.~\ref{fig:jiegou3}}. {According to~\cite{huang2025conditional}, local detail features contain noise that interferes effective feature extraction. To address this issue, we apply {$\text{L}_{1}$ Normalization} $\text{L}_{1}(\cdot)$ for denoising, followed by Softmax Fusion $\text{SMF} (\cdot)$~\cite{li2018densefuse} for feature fusion as below:}
\begin{equation}
\begin{aligned}
&\check{F}^{i}_{L,RGB}, ~\check{F}^{i}_{L,IR} = \text{L}_{1} ({F}^{i}_{L,RGB}), \text{L}_{1} ({F}^{i}_{L,IR}),\\
&\check{F}^{i}_{L}=\text{SMF}(\check{F}^{i}_{L,RGB},\check{F}^{i}_{L,IR}).
\end{aligned}
\tag{9}
\end{equation}
$\check{F}^{i}_{L,RGB}$ and $\check{F}^{i}_{L,IR}$ denote the denoised local features of RGB and IR, respectively. $\check{F}^{i}_{L}$ represents the denoised local feature.
{Following~\cite{cordonnier2020relationship}, which demonstrates the effectiveness of CNNs in perceiving local information, we utilize {three successive $2$D $1\times1$ convolutional layers} to process $\check{F}^{i}_{L}$. The resulting feature is then multiplied by $\check{F}^{i}_{L,RGB}$ refined through spatial attention, and $\check{F}^{i}_{L,RGB}$ refined through channel attention, forming the holistic attention output for $\check{F}^{i}_{L,RGB}$. The attention output is subsequently multiplied with $\check{F}^{i}_{L,RGB}$ to obtain the temporary RGB feature. An identical procedure is applied to $\check{F}^{i}_{L,IR}$, together with the convolutional outputs, to derive the temporary IR feature}. The two temporary features are added to the residual initial features to obtain the final fused local features $\tilde{F}^{i}_{L,RGB}$ and $\tilde{F}^{i}_{L,IR}$ as follows:
\begin{equation}
\begin{aligned}
\tilde{F}^{i}_{L,RGB}=&{F}^{i}_{L,RGB}+\check{F}^{i}_{L,RGB}\odot \text{SA}(\check{F}^{i}_{L,RGB}) \\
&\odot \text{CA}(\check{F}^{i}_{L,RGB}) \odot \text{Conv}(\check{F}^{i}_{L}),\\
\end{aligned}
\tag{10}
\end{equation}
\begin{equation}
\begin{aligned}
\tilde{F}^{i}_{L,IR}=&{F}^{i}_{L,IR}+\check{F}^{i}_{L,IR}\odot \text{SA}(\check{F}^{i}_{L,IR}) \\&\odot \text{CA}(\check{F}^{i}_{L,IR}) \odot \text{Conv}(\check{F}^{i}_{L}).
\end{aligned}
\tag{11}
\end{equation}
$\text{SA} (\cdot)$ denotes Spatial Attention, $\text{CA} (\cdot)$ means Channel Attention, and $\text{Conv}(\cdot)$ represents {the three-layer $2$D $1\times1$ convolutional layers}. As shown in the {bottom} of Fig.~\ref{fig:jiegou3}, {specifically, before being multiplied with the initial features, Spatial Attention separates the features along the {height and width} using average pooling, followed by four DW convolutions, a group {convolution}, and the sigmoid function. }Channel Attention, on the other hand, processes the features through 2D average pooling, a linear layer, ReLU activation, and another linear layer to obtain the final result.

\section{Experiments}
\subsection{Experimental Setups}
\noindent\textbf{Datasets.} We evaluate our LDFE method on {six} public RGB-IR object detection benchmark datasets: M$^3$FD~\cite{liu2022target}, DroneVehicle~\cite{sun2022drone}, LLVIP~\cite{jia2021llvip}, FLIR~\cite{reference1}, KAIST~\cite{hwang2015multispectral}, and VEDAI~\cite{razakarivony2016vehicle}.\par
M$^3$FD is a widely used multi-category RGB-IR detection dataset, consisting of $4,200$ pairs of images. {These images are captured under various extreme conditions, such as low light, heavy rain, snow, and fog, posing significant challenges to the performance of multimodal object detectors.} Since the official dataset partitioning is not provided, we adopt the widely used partitioning method proposed by~\cite{liang2023explicit}, with $80$\% pairs of images used for training and the remaining images for testing.\par
DroneVehicle is a large-scale remote sensing vehicle detection dataset with dense annotations. Its $28,439$ pairs of images with $953,087$ annotations are captured by drones at varying altitudes, angles, and lighting conditions. It contains five categories: car, truck, bus, van, and freight-car. Following the official partitioning standard, we use $17,990$ pairs of images for training, $1,469$ pairs for validation, and $8,980$ pairs for testing, with the test set results used for comparison.\par
LLVIP is a large-scale, well-aligned visible-infrared pedestrian detection dataset. Since the images are captured under low-light conditions, it is widely used for evaluating the performance of multimodal detectors. Following the official partitioning standard, we use $12,025$ pairs of images for training and $3,463$ pairs for testing.\par
FLIR is a publicly available multi-category RGB-IR dataset collected throughout the day. Due to the presence of numerous low-quality images and annotations in the original dataset, we follow~\cite{zhao2024removal} and use FLIR-Aligned dataset, with $4,129$ pairs of images for training and $1,013$ pairs for testing. Due to the limited number of `dog' annotations in the dataset, we removed this category. This dataset is also widely recognized as a challenging RGB-IR object detection benchmark.\par
KAIST is a low-light pedestrian detection dataset similar to LLVIP. Due to problems such as unclear and wrong annotations in the original dataset, we employ the training set ($8963$ image pairs) and {a} test set ($2252$ image pairs) with enhanced annotation quality, which are widely used in the research area.\par
VEDAI is a remote sensing multi-spectral dataset similar to DroneVehicle. It contains $1,210$ pairs of RGB-IR images with annotations for nine different categories. %
Since the dataset does not provide an official split, we use the ten-fold cross-validation method proposed by~\cite{shen2024icafusion}, with $1,089$ image pairs allocated to the training set and the remaining pairs used for testing.\par
\begin{table}[!t]
  \caption{{Comparison results with SOTA methods on M$^3$FD dataset. The best, second, and third results are highlighted in \textcolor{red}{red}, \textcolor{blue}{blue} and \textcolor{orange}{orange}, respectively. }}
  \vspace{5pt}
  \centering
  \scriptsize{\setlength{\tabcolsep}{3pt}   %
  \renewcommand{\arraystretch}{1.0}\begin{tabular}{cc|cc|cccccc}
    \toprule
    Methods & Backbone& $m$AP$_{50}$ & $m$AP & Car & Bus &Lamp &People&Truck&Motorcycle\\
    \midrule
    YOLOv8l-IR~\cite{yolov8_ultralytics}&CSPDarknet53v8&79.5&53.1&90.0&90.9&63.0&82.9&85.9&64.6\\
    YOLOv8l-RGB~\cite{yolov8_ultralytics}&CSPDarknet53v8&80.9&52.5&91.2&92.9&75.3&70.6&86.0&69.6\\
    \midrule
    (ECCV'24) DAMSDet~\cite{guo2024damsdet}&ResNet50&85.3&57.9&93.2&91.9&87.2&81.8&82.9&74.8\\
    (Ours) LDFE&ResNet50&\textcolor{orange}{92.3}&\textcolor{orange}{65.0}&\textcolor{orange}{95.3}&\textcolor{blue}{94.8}&\textcolor{orange}{92.4}&\textcolor{orange}{92.8}&89.2&\textcolor{blue}{89.5}\\
    \midrule
    (IJCV'24) CoCoNet~\cite{liu2024coconet}&CSPDarknet53v5&81.9&55.1&90.5&\textcolor{orange}{94.7}&{91.6}&68.2&72.9&73.4\\
    (TCSVT'24) MMFN~\cite{yang2024multidimensional}&CSPDarknet53v5&86.2&57.4&93.2&92.1&87.6&83.0&87.4&73.7\\
    (CVPR'24) EMMA~\cite{zhao2024equivariant}&CSPDarknet53v5&83.3&55.4&93.8&83.6&87.7&82.9&73.9&77.7\\
    (TIM'24) KCDNet~\cite{wang2024kcdnet}&CSPDarknet53v5&83.5&56.6&91.2&88.8&81.3&83.5&72.0&83.9\\
    (AAAI'25) $FD^2$-Net~\cite{li2025fd2}&CSPDarknet53v5&83.5&-&93.6&82.7&87.8&83.7&73.8&78.1\\
    (TMM'25) Fusion-Mamba~\cite{dong2025fusion}&CSPDarknet53v5&85.0&57.5&91.9&92.8&84.8&80.3&87.1&73.0\\
    (NIPS'24) E2E-MFD~\cite{zhang2024e2e}&CSPDarknet53v5&87.9&58.1&-&-&-&-&-&-\\
   (Ours) LDFE& CSPDarknet53v5&\textcolor{blue}{92.7}&\textcolor{blue}{65.3}&\textcolor{blue}{95.5}&94.4&\textcolor{blue}{92.5}&\textcolor{blue}{92.9}&\textcolor{blue}{89.9}&\textcolor{red}{90.8}\\
    \midrule
    (YAC'24) RI-YOLO~\cite{chen2024rgb}&CSPDarknet53v8&83.6&56.6&91.6&88.5&75.7&84.1&87.8&73.5\\
    (Sensors'24) MRD-YOLO~\cite{sun2024mrd}&CSPDarknet53v8&86.6&59.3&-&-&-&-&-&-\\
    (TCSVT'25) E$I^{2}$Det~\cite{10877920}&CSPDarknet53v8&86.5&55.5&91.5&92.3&84.9&82.9&\textcolor{orange}{89.6}&77.5\\
    (Ours) LDFE& CSPDarknet53v8&\textcolor{red}{93.1}&\textcolor{red}{65.5}&\textcolor{red}{95.6}&\textcolor{red}{95.5}&\textcolor{red}{93.6}&\textcolor{red}{93.1}&\textcolor{red}{91.3}&\textcolor{orange}{89.3}\\
    \bottomrule
  \end{tabular}}
  \label{tab:m3fd}
\end{table}
\begin{table}[!t]
  \caption{{Comparison results with SOTA methods on DroneVehicle dataset. The best results are highlighted in \textcolor{red}{red} and the second and third results are highlighted in \textcolor{blue}{blue} and \textcolor{orange}{orange}, respectively.}}
  \vspace{5pt}
  \centering
  \scriptsize{\setlength{\tabcolsep}{4pt}   %
  \renewcommand{\arraystretch}{1.0}\begin{tabular}{cc|cc|ccccc}
    \toprule
    Methods & Backbone& $m$AP$_{50}$ & $m$AP & Car & Bus &Truck &Freight-car&Van \\
    \midrule
    YOLOv8l-IR~\cite{yolov8_ultralytics}&CSPDarknet53v8&71.9&49.1&93.4&91.9&69.3&53.7&51.1\\
    YOLOv8l-RGB~\cite{yolov8_ultralytics}&CSPDarknet53v8&70.2&48.6&92.5&91.7&68.8&47.8&50.2\\
    \midrule
    (TGRS'24) $C^2$Former-$S^2$ANet~\cite{yuan2024c}&ResNet50&74.2&47.3&90.2&89.8&68.3&64.4&58.5\\
    (Ours) LDFE &ResNet50&\textcolor{orange}{79.4}&\textcolor{orange}{59.9}&94.7&90.2&\textcolor{orange}{79.6}&\textcolor{orange}{68.3}&\textcolor{orange}{64.1}\\
    \midrule
    (CVPR'24) CSOM-ODAF~\cite{chen2024weakly}&CSPDarknet53v5&77.1&50.1&90.1&89.8&75.6&68.2&61.8\\
    (RSL'24) multimodal DINO~\cite{sun2024adaptive}&CSPDarknet53v5& 72.5&50.3&89.5&88.8&75.4&54.3&54.3\\
    (Ours) LDFE &CSPDarknet53v5&\textcolor{blue}{79.5}&\textcolor{blue}{60.1}&94.7&90.0&\textcolor{blue}{80.0}&\textcolor{blue}{68.4}&\textcolor{blue}{64.2}\\
    \midrule
    (RS'24) DAAB+FFPN~\cite{rs16203904}&CSPDarknet53v8&75.2&56.3&-&-&-&-&-\\
    (Sensors'24) IV-YOLO~\cite{tian2024iv}&CSPDarknet53v8&74.6&56.8&\textcolor{orange}{97.2}&\textcolor{orange}{94.3}&65.4&63.1&53.0\\
    (AEJ'24) YOLO-Fusion~\cite{tang2024yolo}&CSPDarknet53v8&72.1&-&\textcolor{red}{98.4}&\textcolor{red}{96.1}&66.0&53.2&46.9\\
    (CAC'24) FFM~\cite{zhou2024vehicle}&CSPDarknet53v8&77.1&-&\textcolor{blue}{98.1}&\textcolor{blue}{94.5}&74.1&64.4&54.5\\
    (Ours) LDFE&CSPDarknet53v8&\textcolor{red}{79.9}&\textcolor{red}{60.5}&94.8&90.1&\textcolor{red}{80.4}&\textcolor{red}{68.8}&\textcolor{red}{65.3}\\
    \bottomrule
  \end{tabular}}
  \label{tab:dronevechile}
  \vspace{-10pt}
\end{table}
\begin{table}[!t]
  \caption{{Comparison results with SOTA methods on LLVIP dataset. The best, second, and third results are highlighted in \textcolor{red}{red}, \textcolor{blue}{blue}, and \textcolor{orange}{orange}, respectively.}}
  \vspace{5pt}
  \centering
  \scriptsize{
  \setlength{\tabcolsep}{22pt}   %
  \renewcommand{\arraystretch}{1.0}
  \begin{tabular}{cc|cc}
    \toprule
    Methods&Backbone&$m$AP$_{50}$ & $m$AP\\
    \midrule
    YOLOv8l-IR~\cite{yolov8_ultralytics}&CSPDarknet53v8&95.2&62.1\\
    YOLOv8l-RGB~\cite{yolov8_ultralytics}&CSPDarknet53v8&91.9&54.0\\
    \midrule
    (2024) RSDet~\cite{zhao2024removal}&ResNet50&95.8&61.3\\
    (Ours) LDFE &ResNet50&\textcolor{orange}{97.3}&\textcolor{orange}{68.2}\\
    \midrule
    (IF’23) DIVFusion~\cite{tang2023divfusion}&CSPDarknet53v5&93.8&58.1\\
    (PR’24) ICAFusion~\cite{shen2024icafusion}&CSPDarknet53v5&95.2&60.1\\
    (RS'24) ACDF-YOLO~\cite{fei2024acdf}&CSPDarknet53v5&96.5&61.3\\
    (PR'25) MGT~\cite{li2025reference}&CSPDarknet53v5&97.0&63.8\\
    (PR'25) HaarFuse~\cite{wang2025haarfuse}&CSPDarknet53v5&96.3&62.8\\
    (Ours) LDFE &CSPDarknet53v5&\textcolor{blue}{97.6}&\textcolor{blue}{68.4}\\
    \midrule
    (SJ'25) FIAFusion~\cite{ding2025fiafusion}&ELANv7&96.2&62.8\\
    (Sensors'24) FAWDet~\cite{zhao2024object} &CSPDarknet53v8&97.1&62.1\\
    (Ours) LDFE & CSPDarknet53v8&\textcolor{red}{97.8}&\textcolor{red}{68.5}\\
    \bottomrule
  \end{tabular}}
  \label{tab:LLVIP}
\end{table}
\begin{table}[!t]
\caption{{Comparison results with SOTA methods on FLIR-Aligned Dataset. The best, second, and third results are highlighted in \textcolor{red}{red}, \textcolor{blue}{blue}, and \textcolor{orange}{orange}, respectively.}}
\vspace{5pt}
\centering
\scriptsize{\setlength{\tabcolsep}{5pt}   %
  \renewcommand{\arraystretch}{1.0}
\begin{tabular}{lcccccc|cc}
\toprule
Methods&Backbone&Precision&Recall&F1&$m$AP & Parameters & Time (ms) \\
\midrule
YOLOv8-IR~\cite{yolov8_ultralytics}&CSPDarknet53v8&75.1&65.3&69.9&38.3&43.7M&22.0\\
YOLOv8-RGB~\cite{yolov8_ultralytics}&CSPDarknet53v8& 71.6&62.2&66.6&28.2&43.7M&22.0\\
\midrule
(RS'24) DAAB+FFPN~\cite{rs16203904}&ResNet50&80.1&73.1&76.4&42.3&176.3M&68.0\\
(Ours) LDFE&ResNet50&\textcolor{orange}{81.9}&\textcolor{orange}{76.1}&\textcolor{orange}{78.9}&\textcolor{orange}{43.8}&187.6M&57.0\\
\midrule
(PR'24) ICAFusion~\cite{shen2024icafusion}&CSPDarknet53v5&-&-&-&41.4&120.2M&\textcolor{orange}{42.0}\\
(PRL'24) CrossFormer~\cite{lee2024crossformer}&CSPDarknet53v5&78.1&72.8&75.4&42.1& 340.0M& 80.0\\
(TNNLS'24) LRAF-Net~\cite{10144688}&CSPDarknet53v5&81.6&75.3&78.3&42.8&197.3M&99.8\\
(TCSVT'24) MMI-Det~\cite{10570450}&CSPDarknet53v5&-&-&-&40.5&207.5M&51.5\\
(Ours) LDFE&CSPDarknet53v5&\textcolor{blue}{82.1}&\textcolor{blue}{76.3}&\textcolor{blue}{79.1}&\textcolor{blue}{44.0}&\textcolor{red}{56.0M}&\textcolor{red}{38.3}\\
\midrule
(GRSL’24) ESSFN~\cite{xu2024enhanced}&CSPDarknet53v8&81.4&73.5&77.2&42.3&\textcolor{blue}{80.2M}&47.0\\
(Ours) LDFE&CSPDarknet53v8&\textcolor{red}{82.9}&\textcolor{red}{76.7}&\textcolor{red}{79.7}&\textcolor{red}{45.1}&\textcolor{orange}{81.3M}&\textcolor{blue}{40.3}\\
\bottomrule
\end{tabular}}
\label{tab:FLIR}
\end{table}
\begin{table}[htbp]
\centering
\begin{minipage}{0.46\linewidth}  %
\centering
\caption{{Comparison results with SOTA methods on KAIST dataset. The best, second, and third results are highlighted in \textcolor{red}{red}, \textcolor{blue}{blue}, and \textcolor{orange}{orange}, respectively.}}
\vspace{5pt}
\scriptsize{\setlength{\tabcolsep}{1.5pt}   
  \renewcommand{\arraystretch}{1.0}\begin{tabular}{cc|cc}
    \toprule
    Methods&Backbone&$m$AP$_{50}$ & $m$AP\\
    \midrule
    YOLOv8l-IR~\cite{yolov8_ultralytics}&CSPDarknet53v8&56.8&22.4\\
    YOLOv8l-RGB~\cite{yolov8_ultralytics}&CSPDarknet53v8&55.3&21.6\\
    \midrule
    (TCSVT'22) CMDet~\cite{sun2022drone}&ResNet50&68.4&28.3\\
    (Ours) LDFE &ResNet50&\textcolor{orange}{78.2}&\textcolor{orange}{36.5}\\
    \midrule
    (RS'22) RISNet~\cite{wang2022improving}&CSPDarknet53v5&72.7&33.1\\
    (PR’24) ICAFusion~\cite{shen2024icafusion}&CSPDarknet53v5&60.3&-\\
    (Ours) LDFE &CSPDarknet53v5&\textcolor{red}{79.1}&\textcolor{blue}{37.0}\\
    \midrule
    (Sensor’24) IV-YOLO~\cite{tian2024iv}&CSPDarknet53v8&75.4&-\\
    (Ours) LDFE & CSPDarknet53v8&\textcolor{blue}{78.8}&\textcolor{red}{37.2}\\
    \bottomrule
  \end{tabular}}
  \label{tab:KAIST}
\end{minipage}
\hfill
\begin{minipage}{0.46\linewidth}  %
\centering
\caption{{Comparison results with SOTA methods on VEDAI dataset. The best, second, and third results are highlighted in \textcolor{red}{red}, \textcolor{blue}{blue}, and \textcolor{orange}{orange}, respectively.}}
\vspace{5pt}
\scriptsize{\setlength{\tabcolsep}{1.0pt}   
  \renewcommand{\arraystretch}{1.0}\begin{tabular}{cc|cc}
    \toprule
    Methods&Backbone&$m$AP$_{50}$ & $m$AP\\
    \midrule
    YOLOv8l-IR~\cite{yolov8_ultralytics}&CSPDarknet53v8&68.4&39.2\\
    YOLOv8l-RGB~\cite{yolov8_ultralytics}&CSPDarknet53v8&64.5&37.9\\
    \midrule
    (2024) DMM~\cite{zhou2024dmm}&ResNet50&75.0&-\\
(Ours) LDFE&ResNet50&\textcolor{orange}{78.4}&\textcolor{orange}{48.8}\\
    \midrule
    (PR'22) MidFusion~\cite{qingyun2022cross}&CSPDarknet53v5&74.9&46.6\\
    (PR'24) ICAFusion~\cite{shen2024icafusion}&CSPDarknet53v5&76.6&44.9\\
    (RS'24) ACDF-YOLO~\cite{fei2024acdf}&CSPDarknet53v5&78.1&47.9\\
    (Ours) LDFE &CSPDarknet53v5&\textcolor{blue}{78.6}&\textcolor{blue}{49.6}\\
    \midrule
     (ESWA'25) MMFDet~\cite{zhao2025differential}&CSPDarknet53v8&77.5&43.5\\
     (Sensors'24) FAWDet~\cite{zhao2024object}&CSPDarknet53v8&69.2&43.7\\
    (Ours) LDFE & CSPDarknet53v8&\textcolor{red}{78.8}&\textcolor{red}{49.9}\\
    \bottomrule
  \end{tabular}}
  \label{tab:VEDAI}
\end{minipage}
\end{table}
\noindent\textbf{Implementation Details.} We conduct all experiments on a single A100 using a dual-stream framework based on the official YOLOv8~\cite{yolov8_ultralytics}. Except for the LDFE module, all other settings of the model are consistent with those of YOLOv5l and YOLOv8l. For all experiments, the batch size is set to $8$ during training and $16$ during testing. The SGD optimizer is used for optimization with a weight decay of $0.001$ and a momentum of $0.9$. For all the datasets, the input image size is $640\times640$ and the initial learning rate is $0.01$ with $\lambda_{coord}=7.5$. For M$^3$FD and DroneVehicle,  the number of training epochs is set to $250$; for LLVIP, it is set to $80$; for FLIR-Aligned, KAIST and VEDAI, it is set to $120$. All data augmentation and other parameters are set to the default values of YOLOv8.\par
\noindent\textbf{Evaluation Metrics.}
For M$^3$FD, DroneVehicle, LLVIP, KAIST, and VEDAI, we evaluate our model using $m$AP$_{50}$ and $m$AP, which are the most commonly used standard metrics for object detection tasks~{\cite{guo2024damsdet}}. The $m$AP$_{50}$ metric represents the mean AP under IoU of $0.50$ and the $m$AP metric represents the mean AP under IoU ranging from $0.50$ to $0.95$ with a stride of $0.05$~\cite{zhao2024removal}. Since FLIR-Aligned dataset is relatively challenging, we replace $m$AP$_{50}$ with precision, recall, and F1 score to comprehensively assess {the model's performance on this dataset {following~\cite{10144688}}.} We also provide the parameters of our model and the average inference time per image pair ($640\times640$) with $10$ runs on A100. %
\subsection{Comparison with SOTA Methods' Results}
We employ three commonly used CNN backbones: ResNet50, CSPDarknet53v5 and CSPDarknet53v8. {CSPDarknet53v5 and CSPDarknet53v8 are the backbone of YOLOv5 and YOLOv8, respectively. For SOTAs, we also compare methods with other backbones, Transformer and ELANv7 (the backbone of YOLOv7).\\} %
\noindent\textbf{M$^3$FD.} The results of our method on M$^3$FD are summarized in Table~\ref{tab:m3fd}. Compared with $11$ SOTA methods, our method using CSPDarknet53v8, CSPDarknet53v5 and ResNet50 backbones achieves the top three positions in both $m$AP$_{50}$ and $m$AP, surpassing the fourth-ranked method by $5.2$\% and $6.2$\%, respectively. From the perspective of sub-categories, our {methods} based on the three backbones {dominate} the top three positions across almost all categories. {Compared to the single-modality baseline model YOLOv8l, our method achieves a significant performance improvement, demonstrating that LDFE effectively integrates complementary information from both the RGB and IR modalities while suppressing noise.} This achievement is attributed to our use of the Laplacian Pyramid to separate and enhance the global features and detailed local information of the two modalities, enabling outstanding multi-category detection performance under extreme weather conditions.\\ %
\noindent\textbf{DroneVehicle.} As shown in Table~\ref{tab:dronevechile}, our method also achieves SOTA performance on the dense remote sensing dataset DroneVehicle. Compared to $7$ multimodal fusion detection methods, our approach ranks first, second, and third, surpassing the fourth-place method by $2.8$\% in $m$AP$_{50}$ and $3.7$\% in $m$AP. Benefiting from our de-noising of the separated high-frequency detail features and attention-based local feature extraction, our method secures the top three positions in similar categories ``Truck'', ``Freight-car'', and ``Van'', which are difficult for other methods to distinguish. Since our method does not include specific designs for densely annotated targets, its performance in the ``Car'' and ``Bus'' categories is inferior to other methods. However, its overall detection performance significantly surpasses theirs. This result indicates that our method exhibits strong generalization capability across different types of RGB-IR detection datasets, demonstrating its robustness and versatility.\\
\noindent\textbf{LLVIP.} We compare our method with other SOTA methods {on the low-light  pedestrian detection dataset LLVIP}, and the results are in Table~\ref{tab:LLVIP}. Our method based on ResNet50, CSPDarknet53v5 and CSPDarknet53v8 achieves $m$AP$_{50}$ of $97.3$\%, $97.6$\% and $97.8$\%, respectively, surpassing the fourth-place result by $0.7$\%. Additionally, our method attains $m$AP scores of $68.2$\%, $68.4$\% and $68.5$\%, exceeding the fourth-place result by $4.7$\%.\\
\noindent\textbf{FLIR-Aligned.} To further demonstrate the effectiveness of our approach, we conduct a comprehensive comparison with other SOTA approaches on FLIR-Aligned across more evaluation metrics. As presented in Table~\ref{tab:FLIR}, our method achieves SOTA performance across four key metrics: Precision, Recall, F1 score, and $m$AP. Moreover, it exhibits lower parameters and reduced inference time compared to other approaches. Under the same backbone CSPDarknet53v5, our method outperforms LRAF-Net with improvements of $0.8$\% and $1.2$\% in F1 score and $m$AP, respectively, while reducing the parameters by $141.3$M and decreasing the inference time by $61.5$ms. CrossFormer and MMI-Det exhibit a similar trend. This improvement is attributed to the design of our fusion module and the linear complexity of Mamba. This result further confirms that our method effectively improves detection performance while simultaneously reducing computational resource consumption.\\
\noindent\textbf{KAIST.} The results of LDFE on KAIST are shown in Table~\ref{tab:KAIST}. Our method, based on the three backbones, secures the top three in terms of $m$AP$_{50}$ and $m$AP, surpassing the fourth-place method by $3.7$\% and $4.1\%$, respectively. This result, combined with the findings on LLVIP, demonstrates the outstanding performance of our method in low-light detection conditions, highlighting the effectiveness of our fusion approach.\\
\noindent\textbf{VEDAI.} As shown in Table~\ref{tab:VEDAI}, although our model is not specifically designed for small object detection in remote sensing, it, based on the three backbones, achieves $m$AP$_{50}$ of $78.4$\%, $78.6$\% and $78.8$\%, surpassing ACDF-YOLO $0.7$\%. Similarly, $m$AP reaches $48.8$\%, $49.6$\% and $49.9$\%, also exceeding it by $2.0$\%.
\begin{figure*}[!t]
  \centering
   \includegraphics[width=0.95\textwidth]{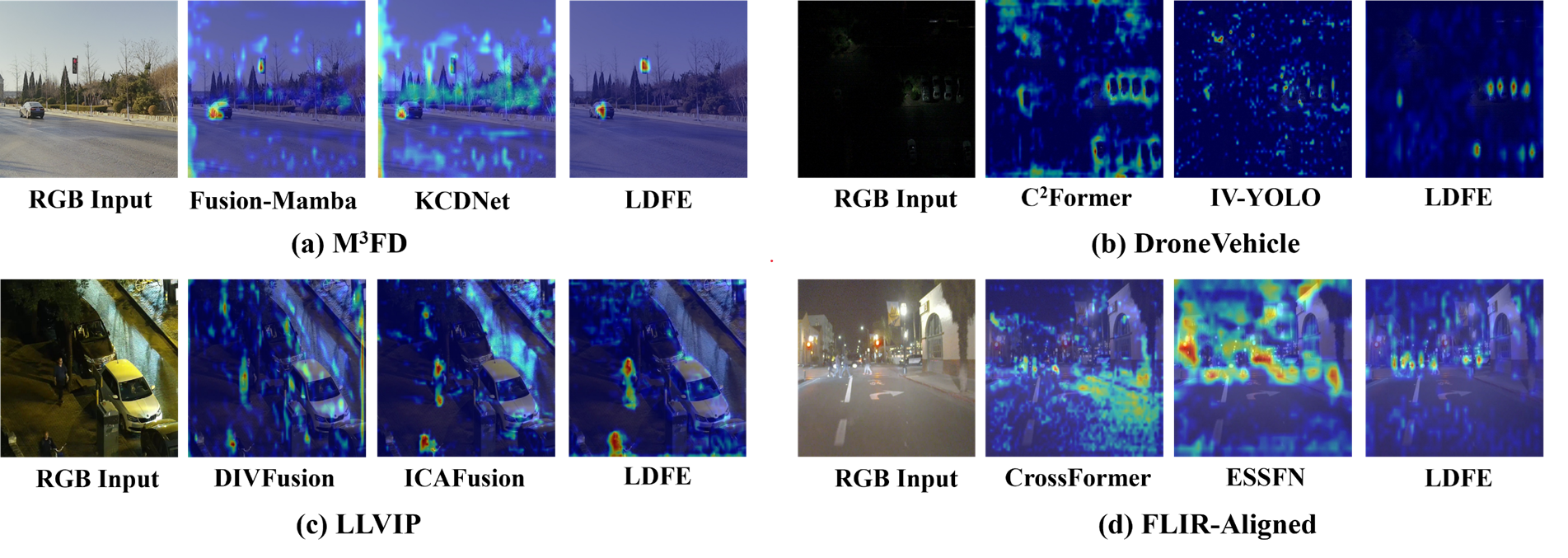}
   \caption{Heatmap visualization of several cross-modal object detection methods on M$^3$FD, DroneVehicle, LLVIP and FLIR-Aligned. Our method focuses more on the target regions.
}
   \label{fig:heatmap}
\end{figure*}
\begin{figure*}[!t]
  \centering
   \includegraphics[width=0.9\linewidth]{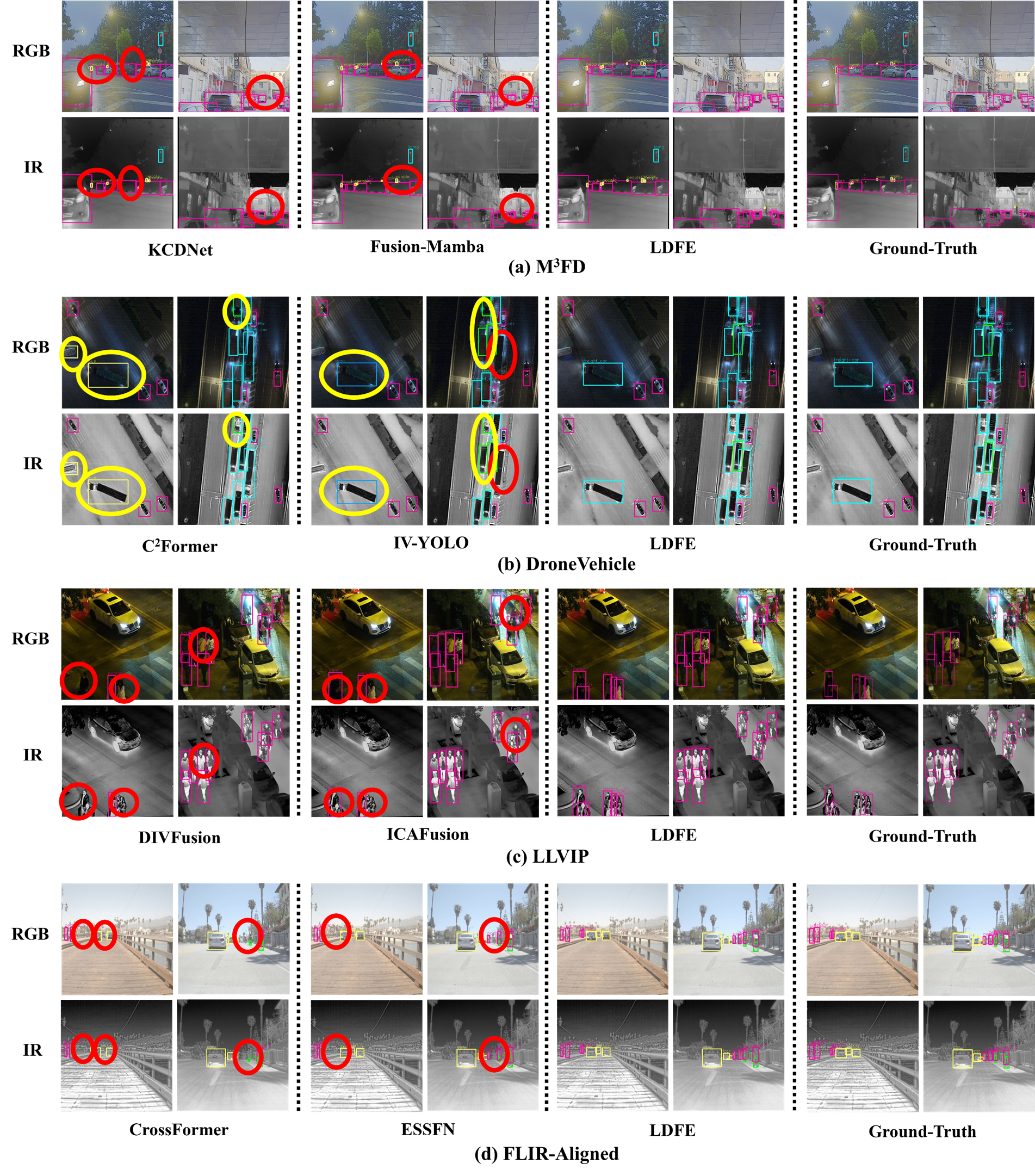}
   \caption{Visualization of detection results of several cross-modal object detection methods on M$^3$FD, DroneVehicle, LLVIP, and FLIR-Aligned. The targets highlighted with red and yellow ellipses represent missed detections and false detections, respectively. Please zoom in for more details.
}
   \label{fig:detect}
\end{figure*}
\subsection{Visualization Results}
Since KAIST and VEDAI are of the same type as LLVIP and DroneVehicle, we provide the visualization results of the latter.\\
\noindent\textbf{Heatmaps.} We visualize the heatmaps of the first LDFE layer based on Grad-CAM~\cite{selvaraju2017grad} and compare them with several {state-of-the-art} methods~\cite{dong2025fusion,wang2024kcdnet,yuan2024c,tian2024iv,tang2023divfusion,shen2024icafusion,lee2024crossformer,xu2024enhanced}. As shown in Fig.~\ref{fig:heatmap}, compared to other methods, our approach focuses more effectively on the target regions while being less distracted by background noise. This is attributed to our method’s denoising and information extraction strategies tailored to {low-frequency global features and high-frequency local features}, along with their associated noise. As a result, the model remains less affected by noise, maintaining focused attention and enhancing detection performance. \\
\noindent\textbf{Detection Results.} We visualize the detection results of our {LDFE} and compare them with several state-of-the-art methods~\cite{dong2025fusion,wang2024kcdnet,yuan2024c,tian2024iv,tang2023divfusion,shen2024icafusion,lee2024crossformer,xu2024enhanced}. As shown in Fig.~\ref{fig:detect}, thanks to our denoising and fusion strategy, our method reduces the number of missed and false detections under challenging conditions (such as far distances, dense annotations, and severe occlusions), achieving the best detection performance.
\begin{table}[!t]
  \caption{{Effects of the CS, GS$^2$E, and LC$^2$E in LDFE on M$^3$FD dataset.}}
  \vspace{5pt}
  \centering
  \scriptsize
  {
  \setlength{\tabcolsep}{18pt}   %
  \renewcommand{\arraystretch}{1.0}
  \begin{tabular}{ccccc}
    \toprule
    Setting&Methods&$m$AP$_{50}$ & $m$AP&Parameters\\
    \midrule
    \multirow{3}{*}{Component}
    &w/o CS &91.6&63.9&81.3M\\
    &w/o GS$^2$E &90.4&61.8&74.2M\\
    &w/o LC$^2$E &90.7&62.1&80.1M\\
    \midrule
    \multirow{5}{*}{Number of LDFEs} 
    &1&91.7&64.2&73.5M\\
    &2&92.1&64.4&75.2M\\
    &3 (Ours)&93.1&65.5&81.3M\\
    \bottomrule
  \end{tabular}}
  \label{tab:ablation1}
\end{table}
\begin{table}[htbp]
\centering
\begin{minipage}[t]{0.45\textwidth}
    \centering
    \caption{{Effects of different fusion methods in LDFE on M$^3$FD dataset. (a,b) means global and local features are fused with methods a and b, respectively.}}
    \vspace{15pt}
    \scriptsize{\setlength{\tabcolsep}{3pt}   %
  \renewcommand{\arraystretch}{1.0}\begin{tabular}{cccc}
    \toprule
    Methods&$m$AP$_{50}$ & $m$AP&Parameters\\
    \midrule
    Baseline&83.2&55.1&73.0M\\
    (Avg, Avg)&86.6&58.2&73.0M\\
    ($L_1$ Norm, $L_1$ Norm)&87.3&58.7&73.0M\\
    (GS$^2$E, Avg)&90.8&62.3&80.1M\\
    (GS$^2$E, $L_1$ Norm)&91.0&62.5&80.1M\\
    (Avg, LC$^2$E)&90.6&62.0&74.2M\\
    ($L_1$ Norm, LC$^2$E)&90.7&62.2&74.2M\\
    (GS$^2$E, LC$^2$E)&93.1&65.5&81.3M\\
    \bottomrule
  \end{tabular}}
  \label{tab:ablation2}
\end{minipage}%
\hfill
\begin{minipage}[t]{0.45\textwidth}
    \centering
    \caption{{Effects of denoising process in LDFE on M$^3$FD dataset.}}
    \vspace{5pt}
    \begin{minipage}[t]{\textwidth}
        \centering
        \scriptsize{\setlength{\tabcolsep}{3pt}   %
  \renewcommand{\arraystretch}{1.0}\begin{tabular}{cccc}
    \toprule
    Methods&$m$AP$_{50}$ & $m$AP&Parameters\\
    \midrule
    Baseline&93.1&65.5&81.3M\\
    w/o denoising in GS$^2$E&91.7&64.3&79.8M\\
    w/o denoising in LC$^2$E&91.9&64.6&81.3M\\
    \bottomrule
  \end{tabular}}
  \label{tab:ablation9}
    \end{minipage}
    
    \vspace{0.05em} %
    
    \begin{minipage}[t]{\textwidth}
        \centering
        \caption{{Effects of different LDFEs' position on M$^3$FD dataset.}}
         \vspace{5pt}
         \scriptsize{\setlength{\tabcolsep}{3pt}   %
  \renewcommand{\arraystretch}{1.0}\begin{tabular}{cccc}
    \toprule
    Methods&$m$AP$_{50}$&$m$AP&Parameters\\
    \midrule
    \{$P_1$,$P_2$,$P_3$\}&91.8&63.2&75.3M\\
    \{$P_2$,$P_3$,$P_4$\}&92.3&64.5&81.4M\\
    \{$P_3$,$P_4$,$P_5$\}&92.1&64.1&87.0M\\
    \{$P_2$,$P_4$,$P_5$\}&92.0&64.0&85.8M\\
    \{$P_2$,$P_3$,$P_5$\} (Ours)&93.1&65.5&81.3M\\
    \bottomrule
  \end{tabular}}
  \label{tab:ablation3}
    \end{minipage}
\end{minipage}
\end{table}
\begin{table}[!t]
  \caption{{Effects of State Space Model on M$^3$FD dataset.}}
  \label{11}
  \vspace{5pt}
  \centering
  \scriptsize
  {
  \setlength{\tabcolsep}{20pt}   
  \renewcommand{\arraystretch}{1.0}
  \begin{tabular}{ccccc}
    \toprule
    Methods&$m$AP$_{50}$&$m$AP&Parameters&Time (ms)\\
    \midrule
    Transformer&91.6&63.7&87.0M&46.7\\
    SSM (Ours)&93.1&65.5&81.3M&40.3\\
    \bottomrule
  \end{tabular}}
\end{table}
\subsection{Ablation Study}
To evaluate the effectiveness of each module, its different configurations, and various fusion strategies, we conduct ablation experiments on M$^3$FD dataset using the CSPDarknet53v8 backbone.\\
\noindent\textbf{Effects of Channel Swapping (CS), GS$^2$E, and LC$^2$E modules.} {As shown in Table~\ref{tab:ablation1}}, we ablate the effect of the CS, GS$^2$E, and LC$^2$E modules by removing them separately. After removing the CS module, the model's performance decreases by $1.5$\% and $1.6$\% in terms of $m$AP$_{50}$ and $m$AP, respectively. Removing GS$^2$E leads to a decrease of $2.7$\% and $3.7$\%, and removing LC$^2$E leads to a decrease of $2.4$\% and $3.4$\% in $m$AP$_{50}$ and $m$AP, respectively. This result validates the importance of each module in our method, demonstrating that the channel swapping, global and local de-noising fusion modules effectively integrate features from both modalities. Furthermore, they enable comprehensive extraction of global information and detailed information from both modalities.\\
\noindent\textbf{Effects of the Number of LDFEs.} 
{Based on the number of detection heads in the YOLOv8 model, we conduct an ablation study on the number of modules. As shown in Table~\ref{tab:ablation1}, using a single fusion module results in decreases of $1.4$\% and $1.3$\% in $m$AP$_{50}$ and $m$AP, respectively, while using two fusion modules leads to decreases of $1.0$\% and $1.1$\% in $m$AP$_{50}$ and $m$AP.}\\
\noindent\textbf{Effects of Different Fusion Methods.} To validate the effectiveness of {our low-frequency global and high-frequency local feature fusion strategies}, we replace them with two standard fusion methods~\cite{li2018densefuse}, Avg Fusion and $L_1$ Norm Fusion. The results are presented in Table~\ref{tab:ablation2}. The baseline model does not utilize Laplacian Pyramid, but directly fuses RGB and IR features using Avg Fusion. Comparing the 1st with the 2nd rows, after incorporating Laplacian Pyramid, the model's performance significantly improves. By comparing the 2nd and 3rd rows with the 4th and 5th rows, we observe that applying GS$^2$E to global features leads to a significant performance improvement while the parameter count increases by only $7.1$M. Similarly, comparing the 2nd and 3rd rows with the 6th and 7th rows also demonstrates the effectiveness of LC$^2$E. This also indicates that using a simple fusion strategy for local features does not adequately capture {the edge, texture, and other detailed information.}\\
\noindent\textbf{Effects of Denoising Process.} {To verify the effect of the denoising process in GS$^2$E and LC$^2$E, we modify GS$^2$E and LC$^2$E by directly passing them through the Mamba module and removing the $L_{1}$ Normalization, respectively. The results are presented in Table~\ref{tab:ablation9}.} After removing the denoising process, the model's $m$AP$_{50}$ and $m$AP show a maximum decrease of $1.4$\%, demonstrating the effectiveness of global and local feature denoising.\\
\noindent\textbf{Effects of Different LDFEs' Position.} Based on previous work~\cite{lee2024crossformer} and the number of detection heads in the YOLOv8 model, we also adopt three fusion modules. The results are summarized in 
Table~\ref{tab:ablation3}, where $P_i$ represents the LDFE is placed at $i^{th}$ {stage}. Due to the results of ``\{$P_1$,$P_2$,$P_3$\}" and~\cite{yuan2024c}, the first {stage} is not a suitable place for fusion. Removing $P_1$ leaves four possible combinations of fusion positions. Ultimately, ``\{$P_2$,$P_3$,$P_5$\}" achieves the best performance.\\
{\noindent\textbf{Effects of State Space Model.} To validate the superiority of using the State Space Model for global feature extraction, we replace SS$2$D in GS$^2$E with the multi-head attention of Transformer, while keeping the rest of the components unchanged. The results are shown in Table~\ref{11}. After adopting Transformer, $m$AP$_{50}$ and $m$AP decrease by $1.5$\% and $1.8$\%, respectively, while the parameters and inference time increased by $5.7$M and $6.4$ms. This is because the State Space Model has linear complexity, whereas Transformer has quadratic complexity.
\begin{table}[!t]
  \caption{{Performance of LDFE on M$^3$FD dataset across different scenarios and the best, second, and third results are highlighted in \textcolor{red}{red}, \textcolor{blue}{blue}, and \textcolor{orange}{orange}, respectively.}}
  \vspace{5pt}
  \centering
  \scriptsize{
  \setlength{\tabcolsep}{10pt}   %
  \renewcommand{\arraystretch}{1.0}
  \begin{tabular}{cc|cccc}
    \toprule
    {Methods}&{Backbone} & Day & Overcast & Night & Challenge\\
    \midrule
    (RAL'23) EAEF~\cite{liang2023explicit}&ResNet50&78.3&78.6&89.5&97.9\\
    (WACV'24) SACBAM~\cite{deevi2024rgb}&EfficientNet&75.9&85.1&92.6&82.7\\
(Ours) LDFE&ResNet50&\textcolor{blue}{90.8}&\textcolor{orange}{91.2}&\textcolor{blue}{97.3}&\textcolor{orange}{99.3}\\
     \midrule
     (CVPR'22) TarDAL~\cite{liu2022target}&CSPDarknet53v5&74.5&74.1&89.3&98.3\\
     (Ours) LDFE&CSPDarknet53v5&\textcolor{blue}{90.8}&\textcolor{blue}{91.4}&\textcolor{orange}{96.9}&\textcolor{blue}{99.5}\\
     \midrule
     (Ours) LDFE&CSPDarknet53v8&\tr{90.9}&\tr{92.8}&\tr{97.4}&\tr{99.6}\\
    \bottomrule
  \end{tabular}}
  \label{tab:Day/Night}
\end{table}
\begin{table}[!t]
\centering
\tiny
\caption{{Performance of LDFE under different severity levels on FLIR-Aligned dataset. The best results are highlighted in bold.}}
\vspace{5pt}
\scriptsize{
\setlength{\tabcolsep}{20pt}   %
\renewcommand{\arraystretch}{1.0}
\begin{tabular}{cccc}
\toprule
Severity level & Methods &Backbones&$m$AP\\
\midrule
\multirow{3}{*}{1} 
& (TITS'23) CMX~\cite{zhang2023cmx} &Transformer&31.8 \\
& (KBS'23) IGT~\cite{chen2023igt} &Transformer&32.8 \\
& (Ours) LDFE & CSPDarknet53v8&\textbf{44.9} \\
\midrule
\multirow{3}{*}{3} 
& (TITS'23) CMX&Transformer  & 30.3 \\
& (KBS'23) IGT &Transformer & 32.0 \\
& (Ours) LDFE & CSPDarknet53v8&\textbf{44.4} \\
\midrule
\multirow{3}{*}{5} 
& (TITS'23) CMX &Transformer& 23.9 \\
& (KBS'23) IGT&Transformer & 28.2 \\
& (Ours) LDFE & CSPDarknet53v8&\textbf{42.7} \\
\bottomrule
\end{tabular}}
\label{tab:guobaoguang}
\end{table}

\subsection{Robustness Test}
To more comprehensively evaluate the generalization and robustness of LDFE, we follow the protocol in~\cite{liang2023explicit} to divide M$^3$FD test set into four scene categories: Day, Night, Overcast, and Challenge, which allows us to assess the generalization capability of our method across diverse environmental conditions. In dynamic scene transitions, the pre-configured camera parameters in autonomous driving systems may lack adaptability, potentially leading to overexposure in visible images. To simulate this, we adopt the approach proposed in~\cite{chen2023igt}, introducing overexposure perturbations to FLIR-Aligned test set by assigning different safety levels, where a higher safety level corresponds to a more severe degree of overexposure. This enables a systematic evaluation of robustness of our method.\\
\textbf{Results of Different Scenes.} We evaluate our method based on three different backbones across four scenarios and the results are presented in Table~\ref{tab:Day/Night}. Compared with other methods, our approach consistently ranks among the top three in $m$AP$_{50}$ across all four scenarios. Specifically, our method outperforms the fourth-ranked approach $12.6$\%, $7.7$\%, $4.8$\%, and $1.3$\% on the four scenes, respectively, which demonstrates the superior generalization and adaptability of our method across diverse environments.\\
\textbf{Results of Different Severity Levels.} The performance of our method under three different severity levels is presented in Table~\ref{tab:guobaoguang}, where our approach consistently achieves the best results. Especially during the transition from level three to level five, as the degree of perturbation intensifies, CMX and IGT decrease by $6.4$\% and $3.8$\%, respectively, whereas our method only decreases by $1.7$\%. These results demonstrate the superior robustness of our approach under conditions of overexposure.

\section{Conclusion}
\label{sec:conclusion}
{In this paper, we propose Laplacian Decoupled Feature Enhancement block (LDFE), an innovative RGB-IR feature fusion paradigm considering both modality and structural properties. It introduces a four-step paradigm including Laplacian
decomposition, denoising, fusion, and Laplacian reconstruction, which is applied for different stages of the
dual-stream CNN backbone. 
Specifically, it explores Laplacian Pyramid for global and local feature decomposition, integrates global features via Global State Space Enhancement module (GS$^2$E) with alternating attention for denoising and state space model for fusion, enhances local features through Local Convolutional Correlation Enhancement module (LC$^2$E) with regularization-based denoising and multi-dimensional attention for fusion, and finally reconstructs the features using Laplacian {pyramid}.
Our method with dual-stream CNN backbones built upon YOLO achieves SOTA performance on {six publicly different RGB-IR object detection datasets} while maintaining low parameters and fast inference speed. {However, our method needs to improve performance under dense small objects and necessitates the development of more task-specific designs to enhance the flexibility of the framework. In the future, we will explore the application of LDFE in more architectures and tasks such as RGB-IR image fusion and segmentation.}}

\bibliographystyle{elsarticle-num}
\bibliography{cas-refs}

\end{document}